\documentclass[conference]{IEEEtran}
\IEEEoverridecommandlockouts
\usepackage{cite}
\usepackage{subcaption}
\usepackage[pdftex]{graphicx}
\usepackage{amsmath}
\usepackage{amssymb}
\usepackage{array}
\usepackage[linesnumbered,ruled,vlined]{algorithm2e}
\usepackage{algpseudocode}
\usepackage{xspace}
\usepackage{multicol}
\usepackage{multirow}
\usepackage{lettrine}
\usepackage{textcomp}
\usepackage{xcolor}
\usepackage{geometry}
\usepackage{booktabs}
\def\BibTeX{{\rm B\kern-.05em{\sc i\kern-.025em b}\kern-.08em
    T\kern-.1667em\lower.7ex\hbox{E}\kern-.125emX}}
\begin{document}

\title{QGAPHEnsemble : Combining Hybrid QLSTM Network Ensemble via Adaptive Weighting for Short Term Weather Forecasting\\
}


\author{
\IEEEauthorblockN{
    Anuvab Sen$^{1}$, 
    Udayon Sen$^{2}$, 
    Mayukhi Paul$^{3}$, 
    Apurba Prasad Padhy$^{4}$, 
    Sujith Sai$^{5}$, 
    Aakash Mallik$^{6}$ \\
    and Chhandak Mallick$^{7}$
}
\IEEEauthorblockA{
    $^{1}$Georgia Institute of Technology, Atlanta, United States \\
    $^{2,3}$Indian Institute of Engineering Science and Technology, Shibpur, Howrah - 711103, India \\
    $^{4}$IIT Roorkee, India \\
    $^{5}$National Institute of Technology, Rourkela, Odisha - 769008, India \\
    $^{6}$National Institute of Technology Karnataka, Surathkal - 575025, India \\
    $^{7}$Jadavpur University, Kolkata - 700032, India \\
    Email: sen.anuvab@gmail.com$^{1}$, 
           udayon.sen@gmail.com$^{2}$, 
           mayukhipaul03@gmail.com$^{3}$, \\
           apurba\_pp@ece.iitr.ac.in$^{4}$, 
           ssujith.sai04@gmail.com$^{5}$, 
           aakashmallik.7777@gmail.com$^{6}$ 
           and chhandak.mallick@icloud.com$^{7}$
}
\vspace{-1 cm}
}

\maketitle

\begin{abstract}
Accurate weather forecasting holds significant importance, serving as a crucial tool for decision-making in various industrial sectors. The limitations of statistical models, assuming independence among data points, highlight the need for advanced methodologies. The correlation between meteorological variables necessitate models capable of capturing complex dependencies. This research highlights the practical efficacy of employing advanced machine learning techniques proposing GenHybQLSTM and BO-QEnsemble architecture based on adaptive weight adjustment strategy. Through comprehensive hyper-parameter optimization using hybrid quantum genetic particle swarm optimisation algorithm and Bayesian Optimization, our model demonstrates a substantial improvement in the accuracy and reliability of meteorological predictions through the assessment of performance metrics such as MSE (Mean Squared Error) and MAPE (Mean Absolute Percentage Prediction Error). The paper highlights the importance of optimized ensemble techniques to improve the performance the given weather forecasting task.
\end{abstract}

\section{Introduction}
Accurate weather forecasting is crucial in various sectors such as agriculture, transport, disaster management and public safety. The precise prediction of meteorological conditions is indispensable for informed decision-making, enabling risk mitigation and resource optimization. Amid technological advancements, researchers and meteorologists are working to improve the precision of weather prediction to furnish more dependable and timely information to relevant entities across different geographical and organizational levels. Recognizing the inherent challenges, meteorological time series data often exhibit complex temporal dependencies, non-linear patterns, and irregularities that conventional models struggle to capture effectively \cite{tempo1}\cite{tempo2}. Traditional models which assume independence among data points, may overlook the intrinsic temporal dependencies present in meteorological phenomena. In the field of meteorology, time series prediction has witnessed notable progress with a diverse array of models. Early efforts predominantly relied on statistical methods such as autoregressive integrated moving average (ARIMA) and seasonal-trend decomposition using LOESS (STL)\cite{ar}\cite{stl}. These models, while providing a foundation for time series analysis, often struggled to capture the complexity of atmospheric phenomena characterized by non-linear trends and seasonality.

In recent years, machine learning models have gained prominence, particularly the application of ensemble techniques like Random Forests and Gradient Boosting Machines\cite{rf1}\cite{rf2}. The ensemble approach uses the collective strength of multiple models to improve predictive accuracy and robustness. Additionally, Support Vector Machines (SVM) and K-Nearest Neighbors (KNN) have been explored for their ability to explain patterns in meteorological time series data\cite{knn}. Recurrent Neural Networks (RNNs) have emerged as a powerful class of models for capturing sequential dependencies. Long Short-Term Memory (LSTM) networks, in particular, have demonstrated success in modeling temporal relationships and adapting to the dynamic nature of atmospheric conditions\cite{rnnlstm1}\cite{rnnlstm2}. Furthermore, hybrid models that combine traditional statistical methods with machine learning algorithms have been proposed to use the strengths of both paradigms\cite{P2}\cite{P3}\cite{P4}. While these models represent significant strides in meteorological time series forecasting, challenges persist in accounting for the spatial and temporal heterogeneity of atmospheric processes. Addressing these challenges requires innovative approaches that uses the potential of emerging technologies. 


This paper proposes two novel optimised quantum hybrid ensemble networks BO-QEnsemble and GenHybQLSTM for the given weather forecasting task. Through extensive experimental studies, we empirically demonstrate the superiority of our proposed models over state-of-the-art (SOTA) counterparts, showcasing enhanced performance across various evaluation metrics, including Mean Absolute Percentage Error (MAPE) and Mean Squared Error (MSE). The presented findings contribute to the field of weather forecasting and establish the efficacy of our novel ensemble networks in surpassing existing benchmarks.



\section{PRELIMINARIES}

This sections provides a concise overview of the fundamental concepts of Quantum Long Short-Term Memory (QLSTM), Particle Swarm Optimisation (PSO) and Bayesian Optimisation (BO) \cite{PSO1}\cite{LSTM1} \cite{BO}\cite{dataset2}. These foundational elements are pivotal components deployed within the proposed ensemble approach.

\subsection{Quantum Long Short-Term Memory (QLSTM)}

QLSTM is the quantum extension of LSTM, a variety of the recurrent neural network (RNN) architecture which serves as one of the most efficient models for handling sequential data, making them the perfect candidate for time-series forecasting tasks\cite{lstme}. Figure 1. illustrates the QLSTM architecture.
\begin{figure}[h]
\centering
\includegraphics[height=2.2in]{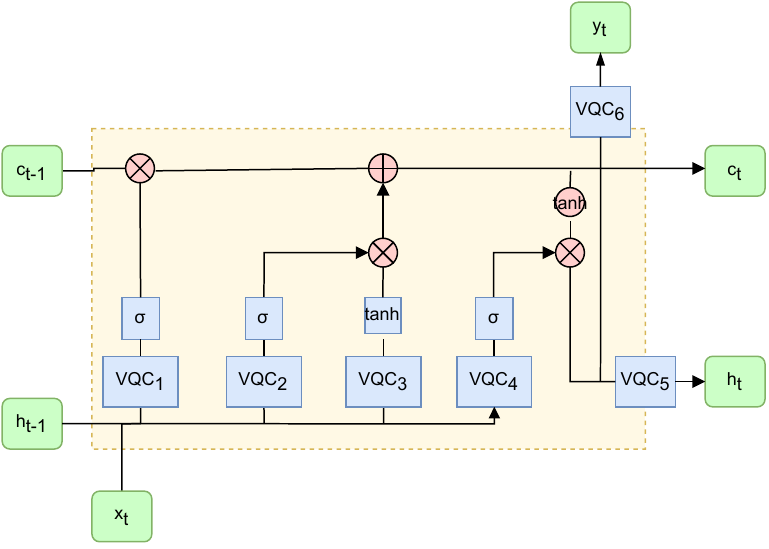}
\label{FS7}
\caption{QLSTM architecture (Variational Quantum Circuits)}
\end{figure}

The mathematical formulation of a QLSTM cell is given as follows:
\begin{equation*}
 f_{t} = \sigma ( VQC_{1}(v_{t} ) 
\end{equation*} 
\begin{equation*}
 i_{t} = \sigma ( VQC_{2}(v_{t} )
\end{equation*}
\begin{equation*}
 \tilde{C}_{t} = tanh(VQC_{3}(v_{t}) )
\end{equation*}
\begin{equation*}
 c_{t} = (f_{t} * c_{t-1} ) + (i_{t} * \tilde{C}_{t} )
\end{equation*}
\begin{equation*}
 o_{t} = \sigma ( VQC_{4}(v_{t})) 
\end{equation*}
\begin{equation*}
 h_{t} = VQC_{5}(o_{t} * tanh(c_{t}))
\end{equation*}
\begin{equation*}
 y_{t} = VQC_{6}(o_{t} * tanh(c_{t}))
\end{equation*}

QLSTM uses Variational Quantum Circuits (VQCs) — quantum circuits with gate parameters optimized (or trained) that replace classical neural networks in LSTM cells, serving as feature extractors and compressors\cite{QLSTM}.
These quantum cells employ quantum entanglement to facilitate better storage and retrieval of sequential information. Every circuit block used in a QLSTM cell consist of three layers: the Forget Block, the Input and Update Block, and the Update Block. To construct a QLSTM cell, the aforementioned VQC blocks are stacked together.

The QLSTM model has 4 variable hyperparameters - 1) Number of epochs, 2) Batch size, 3) Learning rate, and 4) Number of qubits.

\subsection{Particle Swarm Optimisation (PSO)}

Particle Swarm Optimisation is a bio-inspired metaheuristic optimization algorithm based on the behavior of a fish or bird swarm in nature. It aims to search and find the optimal solution in a multidimensional search space by simulating the pattern and movement of particles\cite{pso11}. Like other metaheuristics, PSO begins with the initialization of particles with velocity and position set arbitrarily. Each particle is a representation of a solution. The particle positions and velocities are updated iteratively based on a fitness function and a global best solution found by the swarm. PSO involves the movement of particles through a search space based on mathematical calculations. The update equations for the particle's velocity ($V_{i}$) and position ($X_{i}$) in a typical PSO algorithm are expressed as follows:

The velocity update equation is given by:
\begin{equation}
    V_{i}(t+1) = w \cdot V_{i}(t) + c_{1} \cdot r_{1} \cdot (P_{i}(t) - X_{i}(t)) + c_{2} \cdot r_{2} \cdot (G(t) - X_{i}(t))
\end{equation}

The pseudocode for PSO is given below :

\begin{algorithm}
\caption{Particle Swarm Optimization (PSO)}
Initialize particles' positions and velocities in the search space\;
\For{$iteration = 1$ to $MaxIter$}{
    \For{each particle $i$}{
        Evaluate fitness function $f(x_i)$ for particle $i$\;
        \If{$f(x_i)$ is better than the best fitness value of particle $i$}{
            Update personal best position: $pbest[i] = x_i$\;
            Update personal best fitness value: $pbest\_value[i] = f(x_i)$\;
        }
    }
    Update global best particle: $gbest = $ particle with the best fitness among all particles\;
    \For{each particle $i$}{
        Update particle velocity and position using equations:\;
        $velocity[i] = inertia \times velocity[i] + cognitive\_coefficient \times random() \times (pbest[i] - position[i]) + social\_coefficient \times random() \times (best - position[i])$\;
        $position[i] = position[i] + velocity[i]$\;
    }
}
\Return $gbest$\;
\end{algorithm}

where:
- $V_{i}(t+1)$ is the updated velocity of particle $i$ at time $t+1$,
- $w$ is the inertia weight,
- $c_{1}$ and $c_{2}$ are the cognitive and social coefficients, respectively,
- $r_{1}$ and $r_{2}$ are random values between 0 and 1,
- $P_{i}(t)$ is the best-known position of particle $i$ at time $t$,
- $G(t)$ is the best-known position of the entire swarm at time $t$, and
- $X_{i}(t)$ is the current position of particle $i$ at time $t$.
The position update equation is given by:
\begin{equation}
 X_{i}(t+1) = X_{i}(t) + V_{i}(t+1) 
 \end{equation}

These equations govern the movement of particles in the PSO algorithm, facilitating the exploration and exploitation of the search space to find optimal solutions. Adjusting the parameters like inertia weight ($w$), cognitive and social coefficients ($c_{1}$ and $c_{2}$), and others can impact the algorithm's performance in different optimization scenarios.

\subsection{Bayesian Optimisation (BO)}

Bayesian optimization is an approach to optimizing objective functions that take a long time (minutes or hours) to evaluate. It is best-suited for optimization over continuous domains of less than 20
dimensions, and tolerates stochastic noise in function evaluations. The ability to optimize expensive black-box derivative-free functions makes Bayes Optimisation extremely versatile. Bayes Optimisation consists of two main components: a Bayesian statistical model for modeling the objective
function, and an acquisition function for deciding where to sample next. It builds a surrogate for the objective and quantifies the uncertainty in that surrogate using a Bayesian machine learning technique, Gaussian process regression, and then uses an acquisition function defined from this surrogate to decide where to sample.


\section{Proposed Methodology}

\label{sec:majhead}
\begin{figure*}[t]
\centering
 \includegraphics[width=\textwidth,height=7.5 cm]{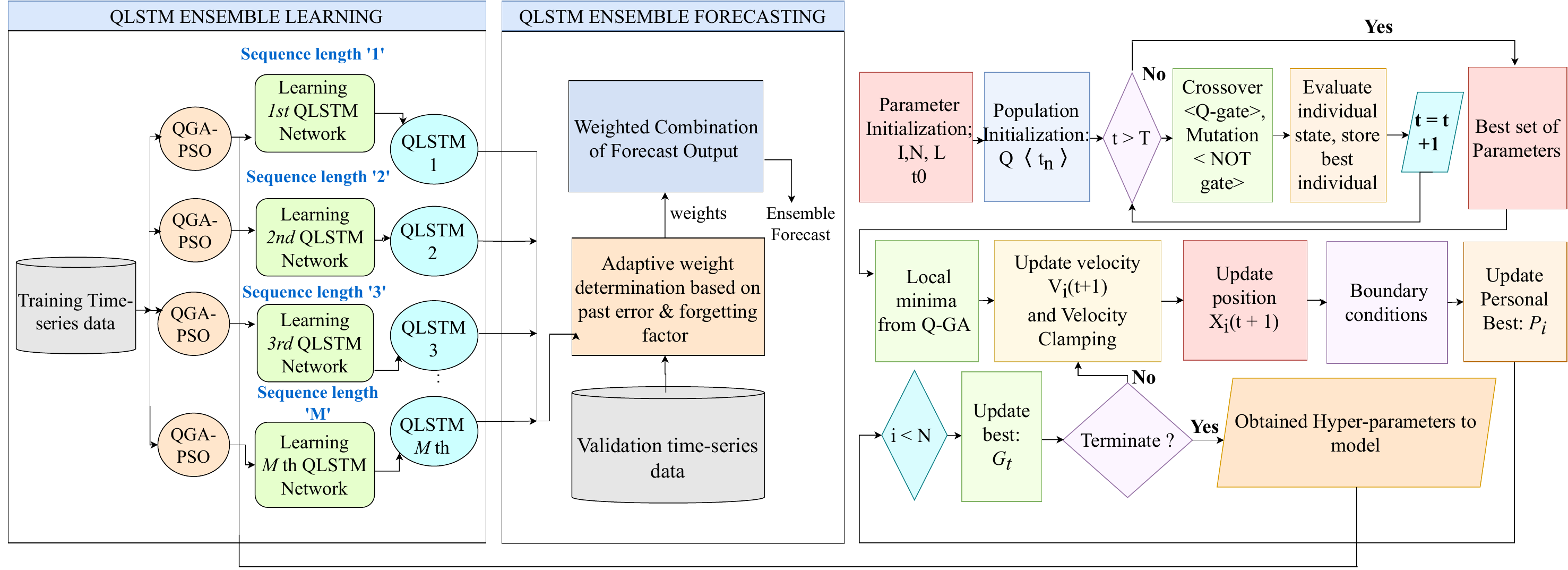}
 \label{FS8}
 \caption{Quantum Genetic-Particle Swarm Algorithm Based QLSTM network ensemble architecture (GenHybQLSTM Ensemble)}
 \end{figure*}

In this section, we explicitly explain the architecture and mechanism of the two ensemble models that we have conceptualized for forecasting 24-hour temperature predictions. Our proposed ensemble learning architecture comprises of two concurrent QLSTMs base models, which are fed with the preprocessed time-series data. The sequence lengths for the QLSTM models are set at $m \in \{3, 5\}$. To optimize the forecasting accuracy of out ensemble model, we have employed a hybrid quantum genetic algorithm - particle swarm optimization approach to fine-tune the hyper-parameter settings before actually training the QLSTM.

In the following subsections, we describe the two distinct approaches employed for QLSTM ensembling.

\subsection {GenHybQLSTM Ensemble}

In our first methodology, the application of the Quantum Genetic Algorithm - Particle Swarm Optimization (QGA-PSO) approach followed by adaptive weight ensemble technique stands as a pivotal element in elevating the performance of our QLSTM models. By employing QGA-PSO to fine-tune hyperparameters before the QLSTM training phase, we capitalize on its unique ability to explore and refine the model's configuration. This dynamic optimization process empowers the QLSTMs to adapt more effectively to the patterns embedded within the time-series data. Consequently, this optimization step significantly enhances the accuracy and resilience of our ensemble forecasting model.

Following experimental studies, we carefully selected optimal values for the learning rate and batch size of each QLSTM through hyper-parameter tuning.

To optimize these hyperparameters effectively, we employ a Quantum Genetic Algorithm (QGA) encapsulating the QLSTM model.
QGA employs the concept of quantum bits and their superposition states which can be represented as follows:- 
\begin{equation}
    \mid\varphi\rangle = \alpha\mid0\rangle + \beta\mid1\rangle 
\end{equation}
where $\alpha$ and $\beta$ are the complex numbers called the probability amplitude of corresponding state of qubit and satisfy the condition $\mid\alpha^2\mid$ + $\mid\beta^2\mid$ = 1.
The QGA uses quantum bits and superposition states, where qubits exist in the states of $\mid0$⟩, $\mid1$⟩, and superposition. Quantum genetic algorithm applies the probability amplitude of qubits to encode chromosome and uses quantum rotating gates to realize chromosomal updated operation\cite{qga11}. The genetic manipulation of quantum genetic algorithm is mainly through acting on the superposition state or entanglement state by the Quantum rotating gates to change the probability amplitude. The Quantum Rotating gate can been defined as given below:
\begin{equation}
U (\theta_{i}) = \begin{bmatrix} cos \theta_{i} & -sin \theta_{i} \\ sin \theta_{i} & cos \theta_{i}\end{bmatrix}  
\end{equation}Probability amplitudes of qubits encode chromosomes, and quantum rotating gates facilitate chromosomal updates.
The pseudocode for QGA is given below:
\begin{algorithm}
\caption{Quantum Genetic Algorithm (QGA)}
Initialize the quantum individuals in the population\;
\For{$iteration = 1$ to $MaxGeneration$}{
    \For{each individual $i$ in popSize}{
        Mutate each individual of population\;
        \For{iteration=1 to genomeLength}{
         Mutate each qubit in an individual's genome\;
        }

        Update the population with the mutation\;
        
    }

    \For{each individual in population}{
    Evaluate fitness using the objective function\;
    
}
    Select the best offsprings for the next generation\;
    
}
Update global fittest individual: $gfittest = $ individual of the particular generation evaluated to be fittest\;
\Return $gfittest$\;
\end{algorithm}
The QGA converges to an optimal set of hyperparameters by minimizing the validation loss, enhancing the QLSTM's ability to capture and study the patterns in the weather dataset. The QGA-optimized hyperparameters serve as the initial solution for further refinement using the Particle Swarm Optimization (PSO) algorithm. PSO, a bio-inspired optimization algorithm, simulates the movement of particles in a multidimensional search space. The velocity and position of particles are iteratively updated based on a fitness function and the global best solution found by the swarm. This iterative process in PSO facilitates the exploration and exploitation of the search space to find the optimal set of hyperparameters.

PSO involves the movement of particles through a search
space based on mathematical calculations. The update equations for the particle’s velocity ( $V_{i}$) and position ($X_{i}$) in a 
The velocity update equation is given by:
\begin{equation}
    V_{i}(t+1) = w \cdot V_{i}(t) + c_{1} \cdot r_{1} \cdot (P_{i}(t) - X_{i}(t)) + c_{2} \cdot r_{2} \cdot (G(t) - X_{i}(t))
\end{equation}
These equations govern the movement of particles in the
PSO algorithm, facilitating the exploration and exploitation
of the search space to find optimal solutions.
\begin{figure}[htbp]
\centering
\includegraphics[width=0.44\textwidth, height=1.28in]{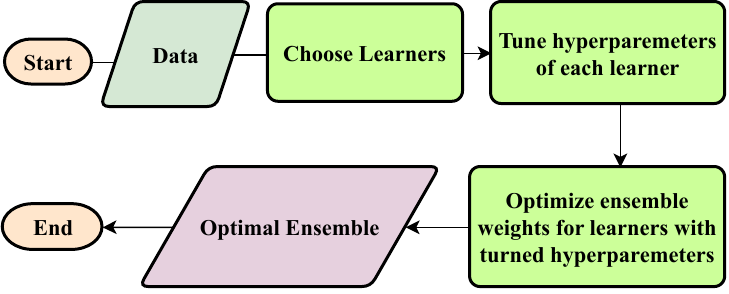}
\label{FS9}
\caption{Flowchart of HybQLSTM Ensemble}
\end{figure}
The hybrid QGA-PSO approach enhances the efficiency of the QLSTM model by utilizing hyperparameters derived from QGA as the initial parameters for PSO, thereby fostering expedited convergence. This collaborative approach yields finely tuned hyperparameters, enhancing the QLSTM model's performance in predicting weather attributes, as demonstrated in subsequent experimental results.

Once the best-suited values for these two critical parameters were identified, we proceeded to train the models and construct the ensemble concurrently.
The aggregation mechanism to combine the predictions from the base QLSTM models is an adaptive weight scheme that accounts for the time varying dynamics of the underlying time series pattern \cite{choi2018combining}. The combining weights are computed by:
\begin{equation}
    w_{m}^{(k+1)}=w_{m}^{(k)}+\lambda\Delta w_{m}^{(k)}\;\;\;\mathrm{for}\;m=1,\ldots,M
\end{equation}

where we have taken $\lambda = 0.85$ .
The $\Delta w_{m}^{(k)}$ is computed based on the inverse prediction error of the respective QLSTM base model as given below:

\begin{equation}
    \Delta w_{m}^{(k)}=\frac{1/\varepsilon_{m}^{(k)}}{1/\varepsilon_{m}^{(1)}+1/\varepsilon_{m}^{(2)}+\cdot\cdot\cdot+1/\varepsilon_{m}^{(k)}}
\end{equation}

The $\varepsilon_{m}^{(k)}$ is related to past prediction error measured up to the th time step in the following way:

\begin{equation}
    \varepsilon_{m}^{(k)}=\sum_{t=k-\nu+1}^{k}\gamma^{k-t}e_{m}^{(t)}
\end{equation}

where $0<\gamma\leq1,1\leq\nu\leq k\ \mathrm{and}\ e_{m}^{(t)}$ is the prediction error at each time step  of the $mth$ QLSTM model. Here, we have taken the forgetting factor $\gamma=0.85$.
\begin{figure*}[t]
     \centering
     \includegraphics[width=0.95\textwidth,height=5.4 cm]{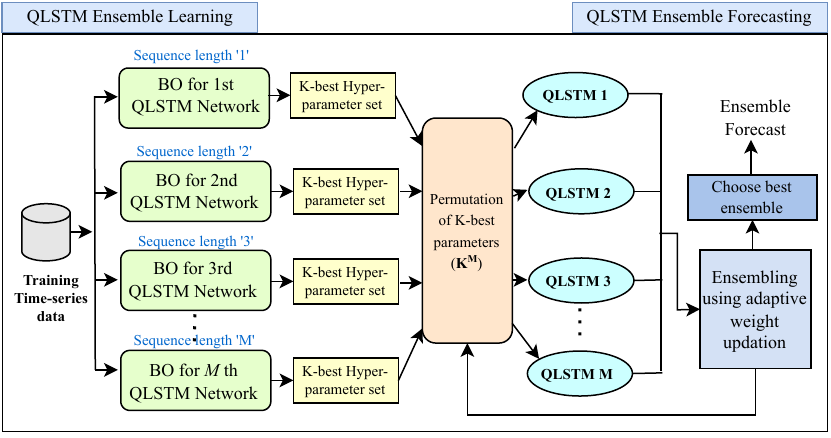}
     \label{FS10}
    \caption{Bayesian based nested optimised QLSTM network ensemble architecture (BO-QEnsemble)}
\end{figure*}
Utilizing the equations (3), (4), and (5), the weights are updated over time series with $T$ time steps $(k= 1,\cdot\cdot\cdot\cdot,M)$ , leading to the final weights $w_{m}^{(T)}(m = 1,\cdot\cdot\cdot\cdot, M)$. Finally, the weight to be assigned to each LSTM model in an ensemble is computed as follows:

\begin{equation}
    w_{m}=\frac{w_{m}^{(l)}}{\sum_{n=1}^{M}w_{n}^{(T)}}\quad\mathrm{for}\ m=1,\dots,M
\end{equation}

the weights computed using (6) satisfy the constraints $0\leq w_{m}\leq1$ and $\textstyle\sum_{m=1}^{M}w_{m}=1.$

Updating weights based on previous errors, with a forgetting factor, gives more importance to recent errors, improving our model's adaptability.

\subsection{BO-QEnsemble Architecture}
In our pursuit of enhancing forecasting accuracy for 24-hour temperature predictions, we introduce a novel ensemble model— the second architecture in our framework. This innovative approach combines the power of Quantum Long Short-Term Memory (QLSTM) base models with the efficiency of Bayesian optimization for hyperparameter tuning. At the heart of this model lies an ensemble comprising \textit{m} QLSTM base models, each coupled with a dedicated bayesian optimizer. The essence of this ensemble lies in the generation of a set of \textit{K}-best hyperparameter configurations (denoted as $K_m$) from each Bayesian optimizer. These configurations collectively form a hyperparameter space, systematically permuted and applied iteratively to refine the succeeding QLSTM base models within the ensemble.

To elaborate on this approach, let's consider the process:

For hyperparameter tuning, each QLSTM base model, distinguished by its sequence length, is equipped with a bespoke Bayesian optimizer. Bayesian optimization (BO) assumes a pivotal role in locating optimal hyperparameters by minimizing the objective function within a specified domain. Each Bayesian optimizer independently produces a set of \textit{K}-best hyperparameter configurations ($K_m$). The amalgamation of these sets forms the complete hyperparameter space through systematic permutations.(\ref{eq:objective_function_minimization}).

\begin{equation}
x^* = \arg\min_{x \in X} f(x)
\label{eq:objective_function_minimization}
\end{equation}

where \(f(x)\) denotes the score to be minimized,  representing the loss function of the respective QLSTM models, \(X\) is the domain of the hyperparameter values, and \(x^*\) is a combination of hyperparameters that produces the lowest value of the score \(f(x)\). 
\begin{figure}[htbp]
\centering
\includegraphics[width=0.46\textwidth, height=1.8in]{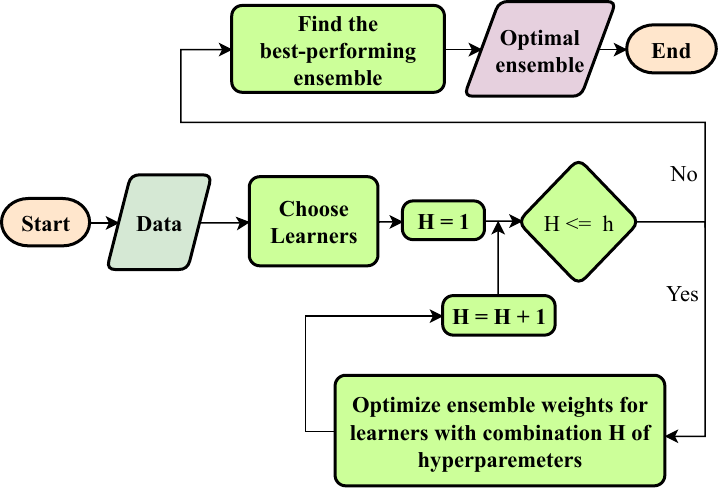}
\label{FS11}
\caption{Flowchart of BO-QEnsemble}
\end{figure}
Bayesian optimizers independently generate a set of K-best hyperparameter configurations ($K_{m}$).
The entire hyperparameter space is formed by permuting these sets. The hyperparameter space is used to fine-tune subsequent QLSTM base models iteratively. 
Predictions from each base model are obtained post-tuning. Then we use the adaptive weighted method for ensemble prediction from these trained QLSTM models. The adaptive weighting equation is defined as follows:
\begin{equation}
    w_{m}^{(k+1)}=w_{m}^{(k)}+\lambda\Delta w_{m}^{(k)}\;\;\;\mathrm{for}\;m=1,\ldots,M
\end{equation}

This method takes into account past errors and incorporates a forgetting factor. 
From the K-best parameter sets generated for \textit{m} models, we can create $K^m$ permutations, resulting in $k^m$ distinct ensemble models. The weight calculation method is kept same as the previous method (equation no. (6), (7), (8) and (9).

The algorithm for the second ensemble model approach is outlined as follows:

\textbf{Inputs:}
\begin{itemize}
    \item Data set $D = \{(x, y): x \in \mathbb{R}^{n \times p}, y \in \mathbb{R}\}$
    \item $m$ base learning algorithm
    \item Hyperparameter sets $h_1, \ldots, h_m$
    \item Bayesian search chooses $k$ hyperparameter combinations for each learner
\end{itemize}

\begin{algorithm}
\caption{Ensemble Model Training with Bayesian Optimization}
\For{$HyperParamCombination = 1, \ldots, b^m$}{
    \For{$model = 1, \ldots, m$}{
        \For{$i = 1, \ldots, n$}{
            Split $D$ into $D_{i,\text{train}}, D_{i,\text{test}}$ for the $i$th split\;
            Train base learner $i$ with hyperparameter combination on $D_{i,\text{train}}$\;
            $P_{i, model}$: Predict on $D_{i,\text{test}}$\;
        }
        $error_{model} = (P_{1,model}, \ldots, P_{m,model})$ 
    }
    Use $error_{model}$ to compute $weights$\;
    Calculate optimal objective value (MSE), optimal weights $(W_{1H}, \ldots, W_{mH})$, and ensemble predictions\;
}
Find the minimum of objective values (MSE)\;
Find the optimal weights $W_{1H}, \ldots, W_{mH}$ corresponding to the minimum objective value\;
\end{algorithm}
\textbf{Outputs:}
\begin{itemize}
    \item Optimal objective value (MSE*)
    \item Optimal combination of hyperparameters $h_1, h_2, \ldots, h_m$
    \item Optimal ensemble weights $w_1, \ldots, w_m$
    \item Prediction vector of ensemble with optimal weights
\end{itemize}

Experimentation with this model architecture has demonstrated a noticeably improved performance compared to the GenHybQLSTM model. This enhancement is likely attributed to the systematic permutation of hyperparameters for each individual model. A detailed study of the results from both models is done in the Results and Discussion section of this paper where we have compared with other state of the art models.


\section{EXPERIMENT DETAILS}

\subsection{Dataset Description}
The dataset utilized in this research was created by scraping from the official website of the Government of Canada weather-related data for the region of Ottawa from January 1st, 2010 to December 31st, 2020 \cite{dataset} \cite{dataset2}. It has the following features: date, time (in 24 hours), temperature, dew point temperature, relative humidity, wind speed, visibility (in kilometers), Pressure (in kilopascals), and precipitation amounts (in millimeters). It also had a few derived features like humidity index and wind chill, which we did not take into account to keep our list of features as independent as possible from each other. The compiled data set comprised 96,432 rows of data for 8 variables, where each row represents an hour.
\subsection{Dataset Preprocessing}
The raw scraped dataset underwent a preprocessing routine wherein the missing data points were imputed using the median values of the respective feature columns. Subsequently the features were scaled using the RobustScaler standardization procedure. In this the median is removed and the data is scaled according to the Inter- Quartile Range (IQR). The IQR is the range between the 1st quartile (25th quantile) and the 3rd quartile (75th quantile). This was done to scale the features in such a way that it is robust to outliers. Following this, 87\% of the 96,432 rows was allocated for training purpose and the remaining 13\% for model testing, to ensure a balanced representation in both the training and testing subsets. Subsequently, Z-score normalization was implemented to ensure a homogeneous distribution of feature scales across the dataset. Following this the values of each feature was transformed such that they have a mean of 0 and a standard deviation of 1.

\subsection{Experimental Setup}

The experiments of this work are implemented in Python 3.10.11 and utilizes the libraries such as Numpy 1.24.3, Pandas 2.0.3, PyTorch 2.1.1 and Matplotlib. Quantum Machine Learning Library - PennyLane 0.34.0 was pivotal in integrating quantum circuits and functionalities into the models.

\section{RESULTS AND DISCUSSION}
In this section, we present the outcomes of our experimental evaluation, analyzing the comparative performance of various models employing distinct optimization algorithms. In the course of our investigation, we have proposed our novel methodologies, GenHybQLSTM and BO-QEnsemble and compared it against other state-of-the-art models. StandardScaler was employed to enhance the convergence and stability in seasonal data training, averting dominant feature emergence. Mean Squared Error serves to minimize training loss. Our primary goal is to enhance forecasting and accelerate convergence, achieved through the integration of meta-heuristics. We introduce a hybrid quantum algorithm for QLSTM hyper-parameter tuning, addressing issues in conventional quantum genetic algorithms, such as premature convergence and limited local search ability. To overcome these limitations without compromising model performance, we incorporate a hybrid QGA-PSO parameter optimization. This approach mitigates premature loss of diversity in the population, often encountered with fitness-driven evaluations, and enhances local search performance, ensuring a more expedited and effective convergence.

Furthermore, the employment of the Bayesian Optimization algorithm within the BO-QEnsemble model, substantiated by commendable results, attests to its efficacy in ascertaining the optimal hyper-parameter setting. Proficient in global optimization, it navigates complex spaces, specifically calibrated for temporal trend prediction. Employing a probabilistic surrogate model, it provides predictions and quantifies uncertainty, enhancing BO-QEnsemble performance in the QLSTM ensemble design.
\begin{table}[htbp]
    \centering
    \caption{Hyperparameter sets and ensemble weights for QLSTM's}
        \begin{tabular}{lcccccc}
            \toprule
            \textbf{Parameter} & \textbf{GenHybQLSTM} & \textbf{BO-QEnsemble}  \\
            \midrule
            \textbf{Learning Rate} & [0.0677, 0.0691] & [0.0726, 0.0522] \\
            \textbf{Layers} & [2, 3] & [2, 2]\\
            \textbf{Number of Qubits} & [6, 6] & [2, 4] \\
            \textbf{Hidden Units} & [5, 5] & [5, 4] \\
            \textbf{Sequence Length} & [3, 5] & [3, 5] \\
            \textbf{Batch size} & [66, 160] & [64,224] \\
            \textbf{Combining Weight} & [0.50272, 0.49728] & [0.50123, 0.49876] \\
            \bottomrule
        \end{tabular}
\end{table}
The choice of Quantum LSTM over classical LSTM as the foundational models for our ensemble architecture significantly contributes to the observed performance enhancement. QLSTM demonstrates accelerated learning, extracting more information immediately after the first training epoch, with a more pronounced and less volatile decrease in both train and test loss compared to the classical LSTM counterpart.
\begin{figure}[htbp]
\centering
\includegraphics[width=0.46\textwidth, height=1.7in]{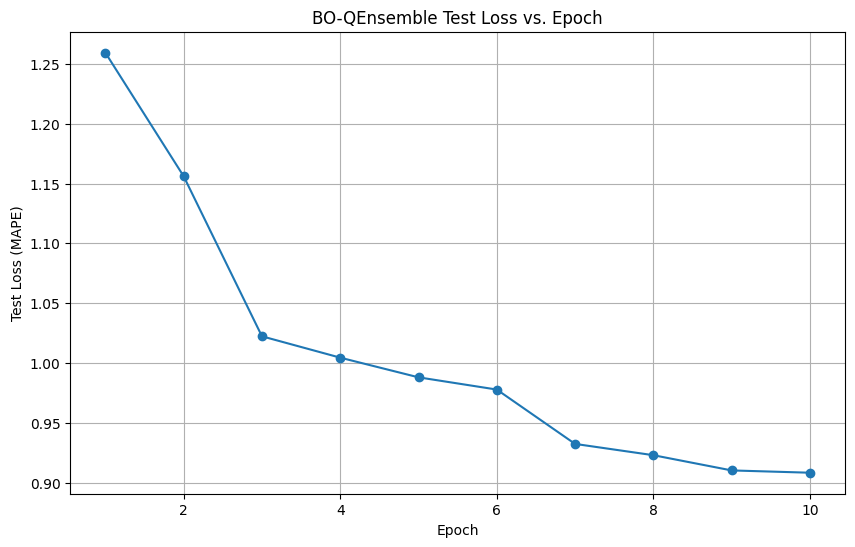}
\label{FS6}
\caption{Test loss Vs Epoch}
\end{figure}
Leveraging quantum circuits modeled after quantum computing phenomena, QLSTM exhibits faster convergence and superior identification of local features, such as minima and maxima, particularly advantageous when confronted with complex temporal structures within the input sequential data.
The test loss vs epoch graph for best performing BO-QEnsemble model is shown in Fig $6$.



Notably, the results underscore the superior efficacy of the BO-QEnsemble, which consistently outperforms alternative models by a substantial margin. Specifically, BO-QEnsemble attains the most favorable outcome with a Mean Absolute Percentage Error (MAPE) of 0.91, surpassing GenHybQLSTM (MAPE: 0.92) and QLSTM (MAPE: 1.12). A comprehensive summary of MAPE of these models is provided in Table $2$. All reported values are aggregated over ten independent statistical trial runs, and the figures presented are denominated in standardized units.
\begin{table}[htbp]
    \centering
    \caption{Comparision of mean absolute percentage error (mape)}
    \begin{tabular}{l c}
        \toprule
        \textbf{Models} & \textbf{MAPE} \\
        \midrule
        Quantum LSTM & 1.12 \\
        LSTM & 1.99 \\ 
        Proposed BO-QEnsemble & 0.91 \\ 
        Proposed GenHybQLSTM & 0.92 \\ 
        DE-ANN & 1.15\\
        Artificial Neural Network & 2.09\\ 
        Gated Recurrent Unit (GRU) & 1.98\\ 
        \bottomrule
    \end{tabular}
\end{table}
Additionally, Fig $7$ presents the forecast plot of the BO-QEnsemble model, showing its predictive insights, while Fig $8$ displays the forecast plot for the GenHybQLSTM model.
\begin{figure}[htbp]
\centering
\includegraphics[width=0.46\textwidth, height=1.5in]{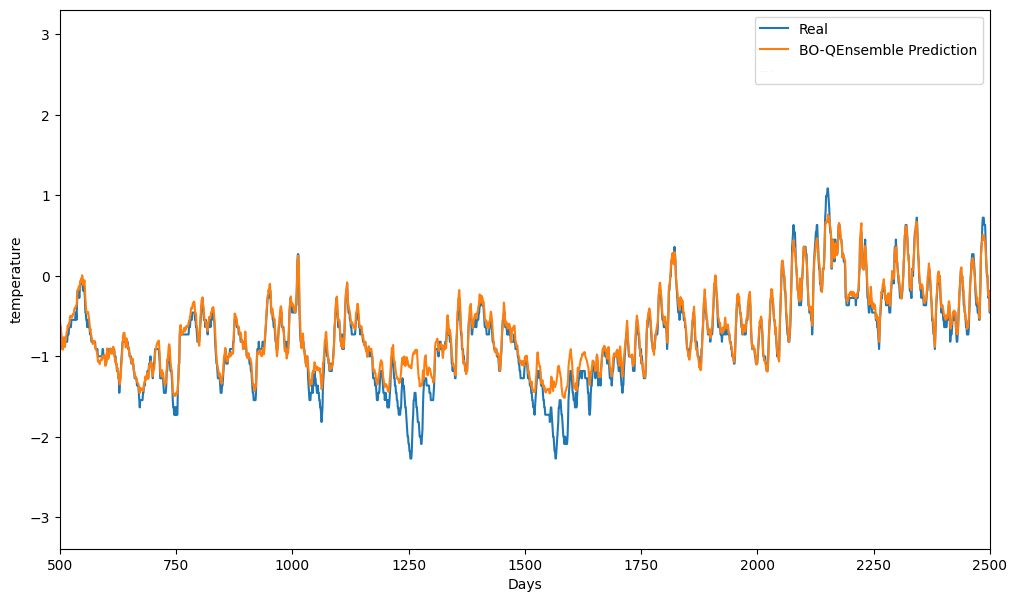}
\label{FS12}
\caption{Forecast Plot for the best performing BO-QEnsemble Model}
\end{figure}
While quantum LSTM shows only a slight improvement in MAPE over DE-ANN, its 
\begin{figure}[htbp]
\centering
\includegraphics[width=0.46\textwidth, height=1.65in]{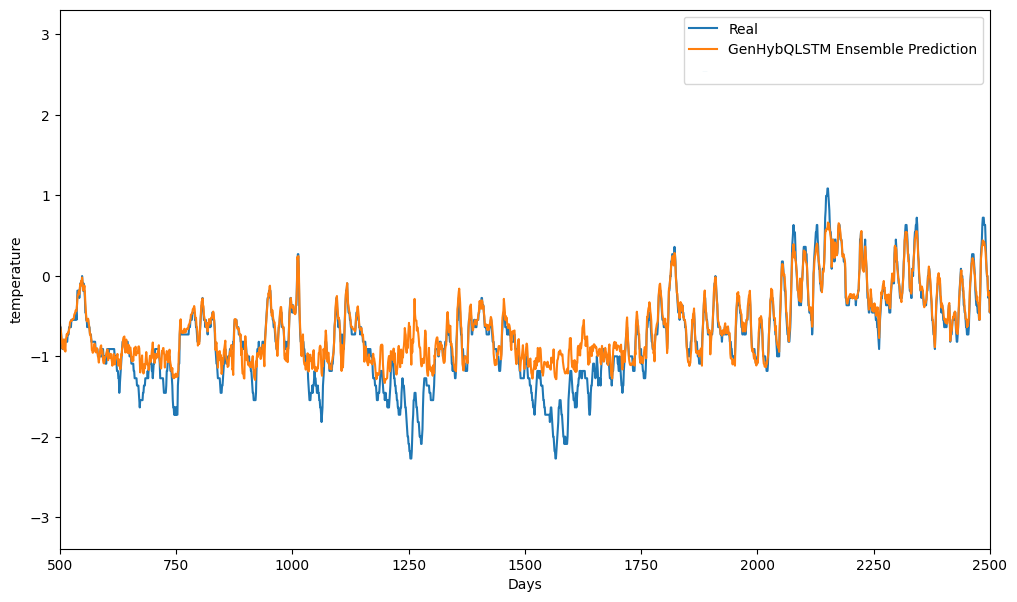}
\label{FS4}
\caption{Forecast Plot for the GenHybQLSTM Ensemble Model}
\end{figure}
computational efficiency outperforms DE-ANN, which is hindered by the complexity of the differential evolution algorithm, resulting in slow performance. Ensemble models consistently outperform individual QLSTM, LSTM, GRU, or ANN models, evident in lower MAPE.
This approach enhances generalization, mitigates overfitting, and improves overall performance on test data by capturing complex temporal patterns during training, reducing variance, and aligning predictions with a realistic range. Figure $9$ shows the 24-hour forecast plot obtained from the best performing BO-QEnsemble model.

\begin{figure}[htbp]
\centering
\includegraphics[width=0.46\textwidth, height=1.6in]{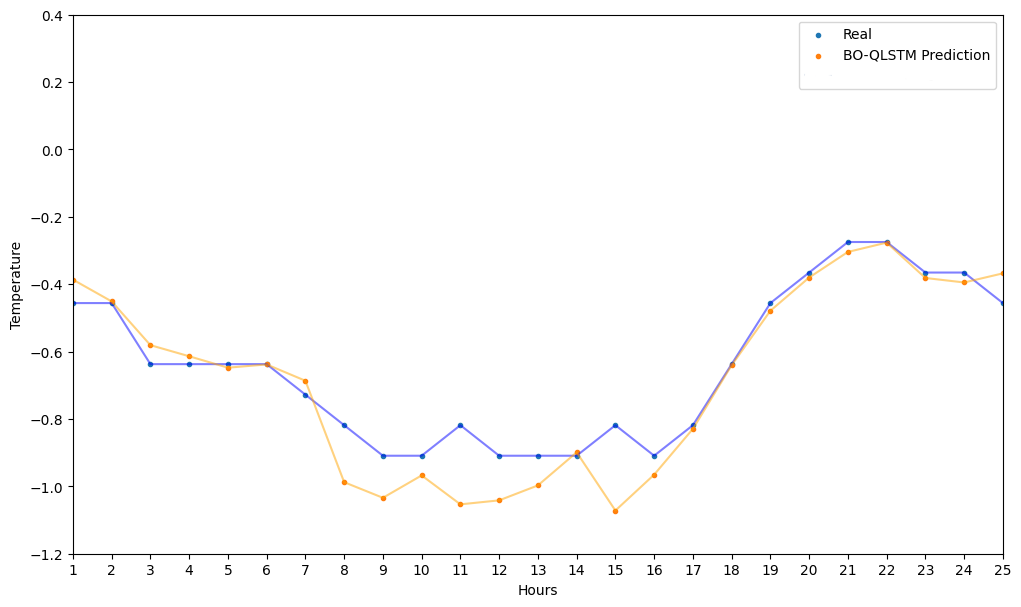}
\label{FS5}
\caption{24-hour Forecast Plot for short term testing of the best performing BO-QEnsemble Model}
\end{figure}


Hence it is conclusive that the proposed ensemble architectures perform better than individual QLSTM networks across a wide range of performance metrics and between the two ensembles the proposed BO-QEnsemble model is superior to the GenHybQLSTM model.

\section{Conclusion and future work}

This manuscript introduces two novel hybrid ensemble models BO-QEnsemble \& GenHybQLSTM arcitectures and a QGA-PSO optimised QLSTM network. These has been compared to other classical deep learning models developed for the forecasting of weather over a speicified geographical area.  Experimental results showcase that our BO-QEnsemble Architecture stands as a SOTA benchmark for the given weather forecasting task and demonstrates superior results in comparision to other models in terms of performance metrics such as mean absolute percentage error (MAPE) and mean squared error (MSE). Future research may be directed towards the incorporation of the proposed ensemble models for testing in different geographical areas and as well as apply it in different application domains.

\vspace{12pt}

\bibliography{strings,refs}
\end{document}